%% file: ML_ijcai13_conj.tex
\documentclass{article}
\usepackage{ijcai13}

\usepackage{times}
\newcommand{\e}{\varepsilon}
\usepackage{ulem}
\usepackage{graphicx, color}
 \usepackage{caption, subcaption}
\usepackage{wrapfig}
\usepackage{amsmath,mathtools, amsfonts, bm}

\usepackage{latexsym} 

\title{Concept Generation in Language Evolution 
}
\author{Martha Lewis, Jonathan Lawry\\
Department of Engineering Mathematics, University of Bristol, BS8 1TR, UK\\
martha.lewis@bristol.ac.uk, j.lawry@bristol.ac.uk}

\begin{document}

\newtheorem{dfn}{Definition}
\newtheorem{lem}[dfn]{Lemma}
\newtheorem{prop}[dfn]{Proposition}
\newtheorem{thm}[dfn]{Theorem}
\newtheorem{cor}[dfn]{Corollary}
\newtheorem {exa}[dfn]{Example}
\newcommand{\vsp}{\vspace{6mm}}
\newcommand{\baralp}[2]{\mbox{$\bar{\alpha}_{#1}^{( #2 )}$}}
\newcommand{\alp}[2]{\mbox{$\alpha_{#1}^{( #2 )}$}}
\newcommand{\vex}[1]{\vec{ #1}}
\newcommand{\V}[1]{\mbox{$V^{(#1)}$}}
\newcommand{\VK}[2]{\mbox{$V^{(#1)}(#2)$}}
\newcommand{\SL}[1]{\mbox{$SL^{(#1)}$}}
\newcommand{\LA}[1]{\mbox{$L^{(#1)}$}}
\newcommand{\fn}{\mbox{$f^{(n)}$}}
\newcommand{\pr}[1]{\mbox{$Prob^{(#1)}$}}

\maketitle
\begin{abstract}
 This thesis investigates the generation of new concepts from combinations of existing concepts as a language evolves. We give a method for combining concepts, and will be investigating the utility of composite concepts in language evolution and thence the utility of concept generation.
\end{abstract}

\section{Introduction}
Humans are skilled at making sense of novel combinations of concepts, so to create artifical languages for implementation in AI systems, we must model this ability. Standard approaches to combining concepts, e.g. fuzzy set theory, have been shown to be inadequate \cite{osh1981}. Composite labels frequently have `emergent attributes' \cite{hamp1987} which cannot be explicated by decomposing the label into its constituent parts. We argue that in this case a new concept is generated. This project aims to determine conditions for such concept generation, using multi-agent models of language evolution.

\subsection{Thesis Outline}
The project divides into three parts. Firstly, we have developed a model of concept combination within the label semantics framework as given in \cite{lawry2009}. The model is inspired by and reflects results in \cite{hamp1987}, in which membership in a composite concept can be rendered as the weighted sum of memberships in individual concepts. 

Secondly, we must show that compositionality can evolve within a population of interacting agents. Preliminary work in this area examines the ability of a population of agents to converge to a shared set of dimension weights.

Thirdly, we will investigate the generation of new unitary concepts from existing composite concepts, building further upon the multi-agent model. 
  
\section{Background}
This work is based on the label semantics framework \cite{lawry2004,lawry2009}, together with prototype theory \cite{rosch}, where membership in a concept is based on proximity to a prototype, and conceptual spaces \cite{gard2004}. The latter views concepts as regions of a space made up of quality dimensions and equipped with a distance metric, for example the RGB colour space.

Label semantics proposes that agents use a set of labels $LA = \{L_1, ..., L_n\}$ to describe a conceptual space $\Omega$ with distance metric $d(x, y)$. Labels $L_i$ are associated with prototypes $P_i \subseteq \Omega$ and uncertain thresholds $\e_i$, drawn from probability distributions $\delta_{\e_i}$. The threshold $\e_i$ captures the notion that an element $x \in \Omega$ is sufficiently close to $P_i$  to be labelled $L_i$. The appropriateness of a label $L_i$ to describe $x$ is quantified by $\mu_{L_i}(x)$, given by 

\[
\mu_{L_i}(x) = P(d(x, P_i) \leq \e_i) = \int_{d(x, P_i)}^\infty  \delta_{\e_i}(\e_i)\mathrm{d}\e_i 
\]

Labels can then be described as $L_i = <\!\!P_i, d(x, y), \delta_{\e_i}\!\!>$.

\section{A New Model of Concept Composition}
\label{sec:cmp}
\begin{figure}[htbp]
\tiny
  \centering  
  \def\svgwidth{0.5\columnwidth}  
  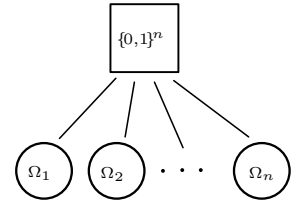  
\caption{Combining labels in a binary space}
\label{fig:binspace}
\end{figure}

Experiments in \cite{hamp1987} propose that human concept combination can (roughly) be modelled as a weighted sum of attributes such as `has feathers', `talks' (for the concept `Bird'). These attributes differ from quality dimensions in conceptual spaces: they tend to be binary, complex, and multidimensional. We therefore view each attribute as a label in a conceptual space $\Omega_i$ and combine these labels in a binary space $\{0,1\}^n$ illustrated in figure \ref{fig:binspace}, where a conjunction of such labels $\tilde{\alpha} = \bigwedge_{i = 1}^n\pm L_i$ maps to a binary vector $\vec{x}_\alpha$ taking value $1$ for positive labels $L_i$ and $0$ for negated labels $\neg L_i$. We treat membership in $\tilde{\alpha}$ in the binary space within the label semantics framework. So $\tilde{\alpha}$ is described in the binary space by $\tilde{\alpha}=<\!\!\vec{x}_\alpha, d(\vec{x}, \vec{x}'), \delta\!\!>$ as before.

We define  a distance metric in the binary space $\{0,1\}^n$ as:
\begin{dfn}{Weighted Hamming Distance}\\
For $\vec{\lambda} \in (\mathbb{R}^+)^n$, $\forall \vec{x},\vec{x}^\prime \in \{0,1\}^n$, where $(\cdot)$ is the scalar product, 
\[
H_{\vec{\lambda}}(\vec{x},\vec{x}^\prime)= \vec{\lambda}\cdot|\vec{x}-\vec{x}^\prime|
\]
\end{dfn}

\begin{thm}
\label{sumthm}
Let $\alpha=\bigwedge_{i=1}^n \pm L_i$ and $\lambda_T = \sum_{i = 1}^n \lambda_i$. Let $\e \sim U(0,\lambda_T)$, $d = H_{\vec{\lambda}}$. Then:
\begin{gather*}
\mu_{\tilde{\alpha}}(\vec{Y})=\sum_{i=1}^n \frac{\lambda_i}{\lambda_T} \mu_{\pm L_i}(Y_i)
\end{gather*}
\end{thm}

Compound concepts $\tilde{\theta}, \tilde{\varphi}$ may be combined in a higher level binary space.
Then $ \tilde{\theta}\bullet \tilde{\varphi}$ can be expressed in the continuous space as a weighted sum of $\tilde{\theta}$ and $\tilde{\varphi}$.
\begin{thm}
\label{conjthm}
 Let $\tilde{\theta} \bullet \tilde{\varphi} = <\!\!\{(1,1)\}, H_{\vec{w}}, \delta\!\!>$. Then $\mu_{\tilde{\theta}\bullet\tilde{\varphi}}(\vec{Y}) = \sum_{i=1}^n (\frac{w_1 \lambda_{\varphi_T} \lambda_{\theta_i} + w_2 \lambda_{\theta_T} \lambda_{\varphi_i}}{w_T \lambda_{\theta_T} \lambda_{\varphi_T}}) \mu_{\pm L_i}(\vec{Y})$.
\end{thm}

We have therefore shown that combining labels in a weighted binary space leads naturally to the creation of composite and compound concepts as weighted sums of individual labels, reflecting results in \cite{hamp1987}. We have further characterised notions of necessary and impossible attributes using ideas from possibility theory.

\section{Convergence of Dimension Weights Across a Population}
\label{sec:conv}
We investigate how a population of agents in a multi-agent simulation playing a series of language games might converge to a shared set of dimension weights. Agents with equal labels $L_1 = L_2 = <\!\!1, d, U[0,1]\!\!>\in \Omega_1 = \Omega_2 = [0, 1]$ ($d$ is Euclidean distance), and randomly initiated weights $\lambda \in [0,1]$ engage in a series of dialogues about elements in the conceptual space, adjusting their weights after each dialogue is completed. At each timestep, speaker agents make assertions $\alpha_i = \pm L_1 \wedge \pm L_2$  about elements $\vec{x} \in \Omega_1 \times \Omega_2$ which maximise $\mu_{\alpha_i}(\vec{x}) = \lambda \mu_{L_1}(x_1) + (1 - \lambda) \mu_{L_2}(x_2)$.

The listener agent assesses $\alpha_i$ against its own label set. If $\mu_{\alpha_i}(x) \leq w$, the reliability of the speaker agent, the listener agent updates its label set.

The update consists in incrementing the dimension weight $\lambda$ towards a value $A$, so that $\lambda_{t+1} = \lambda_t + h(A - \lambda_t)$ where $h = 10^{-3}$ and
\[
A = \frac{w - \mu_{L_2}(x_2)}{\mu_{\pm L_1}(x_1) - \mu_{\pm L_2}(x_2)}
\]

This is the quantity that satisfies $\mu_{\alpha_i}(x) = w$. If $A < 0$ (or $A > 1$) we set $A = 0$ (or $A = 1$).

The convergence across the population is measured by the standard deviation (SD) of the $\lambda$ across the population. 

\begin{figure} [htbp]
         \centering
         \begin{subfigure}[t]{0.2\textwidth}
                 \centering
                 \includegraphics[width=\textwidth]{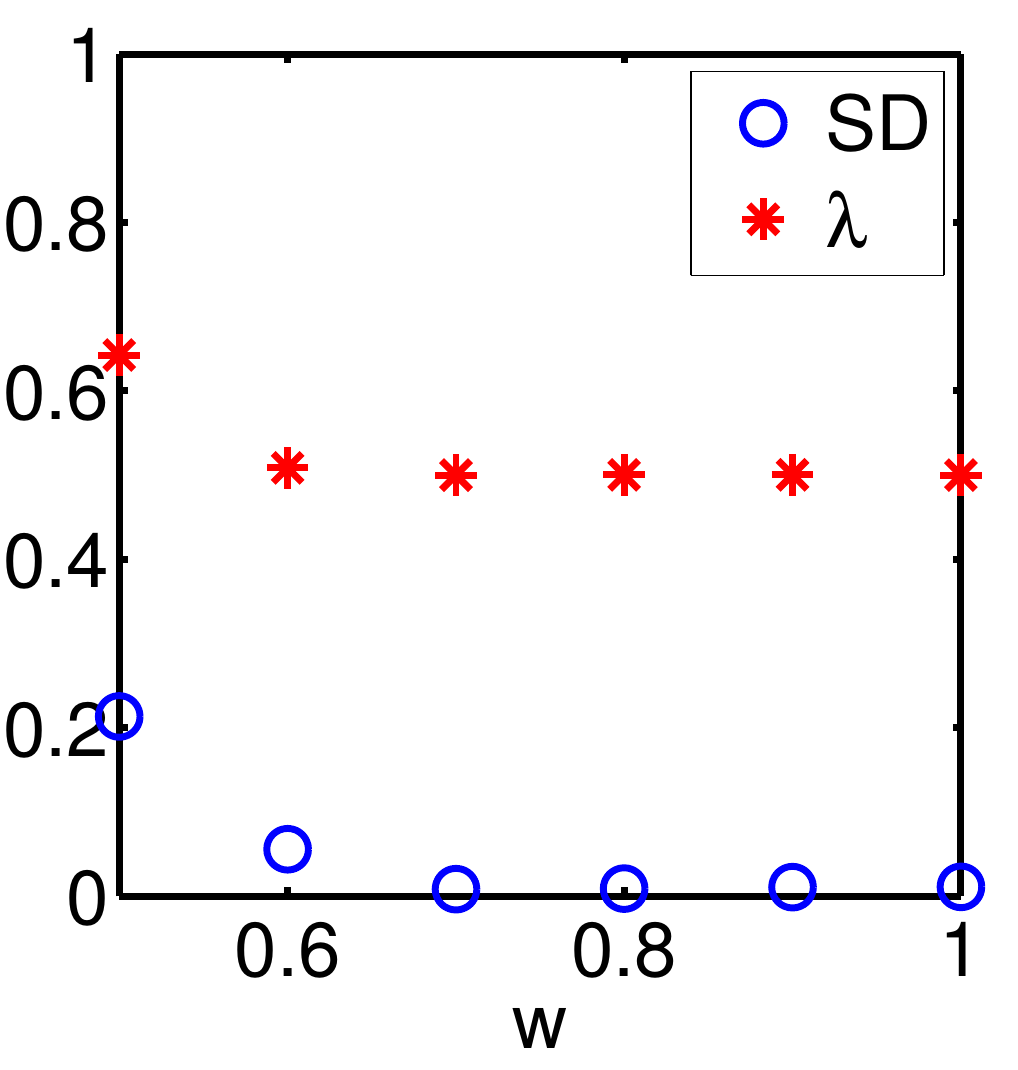}
                 \caption{$x_1 \sim U[0,1]$, $x_2 \sim U[0,0.5]$. $\lambda$ converges to 0.5 for all values of $w$}
                 \label{fig:cmpeq1}
         \end{subfigure}
         ~ 
         \begin{subfigure}[t]{0.2\textwidth}
                 \centering
                 \includegraphics[width=\textwidth]{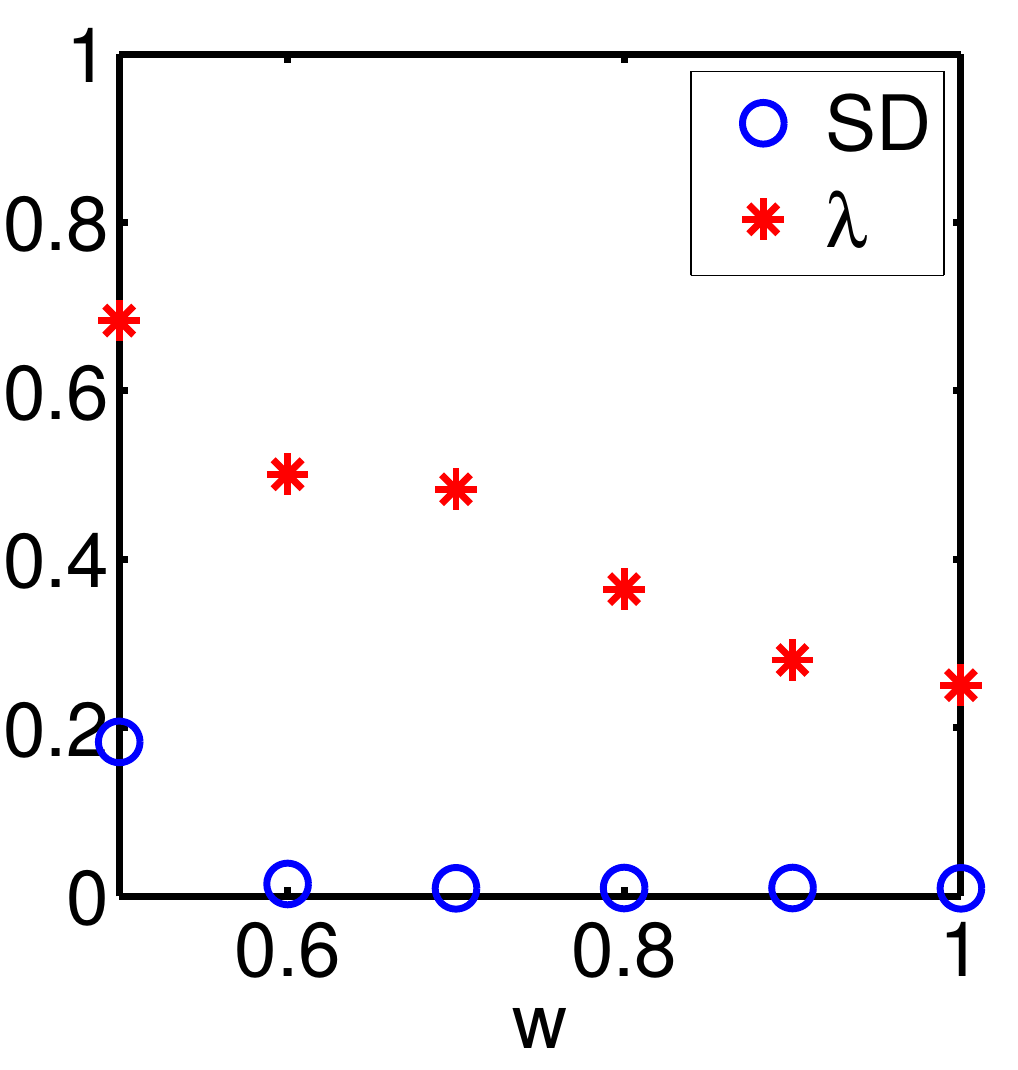}
                 \caption{$x_1 \sim U[0.25,0.75]$, $x_2 \sim U[0,0.5]$. $\lambda$ converges to varying values.}
                 \label{fig:cmpeq075}
         \end{subfigure}
 	\caption{Mean SD and $\lambda$ at time $t = 2000$ for different values of $w$. Each point averages 25 simulations run with 10 agents.}
 	\label{fig:cmpeq}
 \end{figure}

Figure \ref{fig:cmpeq} shows the results of two sets of simulations across varying values of $w$. The two sets of simulations have distinct distributions of elements encountered within the space.  When $w$ is 0.5 or below, the agents do not converge to shared dimension weights (not shown). When $w > 0.5$, agents do converge to shared dimension weights: SD is low. The weights converged to depend both on the reliability, $w$, of each agent, and the distribution of elements in the conceptual space.

When $w = 1$ we can predict the value to which $\lambda$ will converge. Consider the quantity $A-\lambda_t$ which determines whether the update is positive or negative at each step.
\begin{dfn}
A \emph{positive region} $R^+ \subset \Omega$ is a set of points $R^+ =\{\vec{x} \in \Omega: A - \lambda_t \geq 0\}$
\end{dfn}

\begin{thm}
Let $p^+$ denote the probability of a point $\vec{x} \in \Omega$ falling in a positive region and let $w = 1$ across the population. Then the expected value of $\lambda$ converges to $p^+$.
\end{thm}

\section{Further Work}
We are currently working on analytical results to predict the value of $\lambda$ to which agents converge. Under certain circumstances, such as the case where $w = 1$, or with an altered updating model, analytic results are possible. We will extend this work to look at the utility of using conjunctive assertions within these simulations.

Work in the third year will focus on examining how new concepts might be generated from the combination of existing ones. We will build on the language evolution model currently in development.

\bibliographystyle{named}
\bibliography{ijcai13}

\end{document}

%% file: binspace.pdf_tex

\begingroup
  \makeatletter
  \providecommand\color[2][]{%
    \errmessage{(Inkscape) Color is used for the text in Inkscape, but the package 'color.sty' is not loaded}
    \renewcommand\color[2][]{}%
  }
  \providecommand\transparent[1]{%
    \errmessage{(Inkscape) Transparency is used (non-zero) for the text in Inkscape, but the package 'transparent.sty' is not loaded}
    \renewcommand\transparent[1]{}%
  }
  \providecommand\rotatebox[2]{#2}
  \ifx\svgwidth\undefined
    \setlength{\unitlength}{334.0328125pt}
  \else
    \setlength{\unitlength}{\svgwidth}
  \fi
  \global\let\svgwidth\undefined
  \makeatother
  \begin{picture}(1,0.61013471)%
    \put(0,0){\includegraphics[width=\unitlength]{binspace.pdf}}%
    \put(0.03951708,0.07068244){\color[rgb]{0,0,0}\makebox(0,0)[lb]{\smash{$\Omega_1$}}}%
    \put(0.26523405,0.06897508){\color[rgb]{0,0,0}\makebox(0,0)[lb]{\smash{$\Omega_2$}}}%
    \put(0.72506911,0.06897508){\color[rgb]{0,0,0}\makebox(0,0)[lb]{\smash{$\Omega_n$}}}%
    \put(0.31493912,0.48261802){\color[rgb]{0,0,0}\makebox(0,0)[lb]{\smash{$\{\!0,\!1\!\}\!^n$}}}%
  \end{picture}%
\endgroup